\documentclass[sigconf]{acmart}
\AtBeginDocument{%
  }

\usepackage{multirow}

\copyrightyear{2026}
\acmYear{2026}
\setcopyright{cc}
\setcctype{by}
\acmConference[MM '26]{Proceedings of the 34th ACM International Conference on Multimedia}{November 10--14, 2026}{Rio de Janeiro, Brazil}
\acmBooktitle{Proceedings of the 34th ACM International Conference on Multimedia (MM '26), November 10--14, 2026, Rio de Janeiro, Brazil}
\acmDOI{10.1145/3767308.3835757}
\acmISBN{979-8-4007-2213-4/2026/11}

\begin{document}

\title{Path2ST: Hierarchical Cell-Tissue Grounded Cross-Modal Translation for Spatial Transcriptomics}

\author{Ruochen Liu}
\email{sgrliu18@liverpool.ac.uk}
\orcid{0009-0008-5112-7489}
\affiliation{%
  \department{Faculty of Science and Engineering}
  \institution{University of Liverpool}
  \city{Liverpool}
  \country{United Kingdom}
}

\author{Wei Lou}
\correspondingauthor
\email{louwei@zjnu.edu.cn}
\orcid{0000-0002-2071-4081}
\affiliation{%
  \department{College of Mathematical Medicine}
  \institution{Zhejiang Normal University}
  \city{Jinhua}
  \country{China}
}

\renewcommand{\shortauthors}{Ruochen Liu and Wei Lou}

\begin{abstract}
Predicting spatial gene expression from hematoxylin and eosin (H\&E)-stained images offers a cost-effective alternative to spatial transcriptomics (ST). However, existing methods treat H\&E images as generic visual inputs and ignore their intrinsic biological hierarchy, where spatially organized cell types collectively form functional tissue microenvironments that govern local gene expression programs. To bridge this gap, we formulate H\&E-to-ST prediction as a cross-modal semantic translation task and propose Path2ST, a hierarchically grounded autoregressive framework featuring three key components: (i) a Hierarchical Cell-Tissue Conditioning mechanism that fuses explicit and implicit cellular features with tissue-level semantic representations to construct hierarchical conditioning signals; (ii) a Scale-Adaptive Autoregressive Generation process over a hierarchical semantic vocabulary, enabling coarse-to-fine, biologically consistent expression synthesis; and (iii) SpectraLoss, a full-spectrum objective that jointly enforces ordinal fidelity, models transcriptional bursts, and aligns semantic structures with cell types. Extensive experiments on three datasets demonstrate state-of-the-art performance, validating that Path2ST generates highly accurate and spatially coherent transcriptomic profiles. The related code is released at https://github.com/RuochenLiu23/Path2ST.
\end{abstract}



\maketitle

\section{Introduction}
Spatial transcriptomics (ST) enables genome-wide measurement of gene expression while preserving the spatial context of intact tissues, providing unprecedented insight into the molecular organization of complex biological systems~\cite{jain2024review,williams2022introduction,staahl2016st,moses2022museum}. However, its routine application in clinical practice and large-scale studies remains limited by high costs, specialized instrumentation, and labor-intensive workflows~\cite{schroeder2025scaling,rao2021spatial,hu2025scalablest}. By comparison, hematoxylin and eosin (H\&E)-stained whole-slide images are inexpensive, widely available, and already embedded in standard pathology pipelines, often serving as complementary references in ST experiments~\cite{wang2025benchmarking,chelebian2025combining,HistoGene}. While H\&E images do not provide direct molecular readouts, their morphological patterns may encode latent signals associated with local gene expression. This creates an opportunity for AI models to infer spatial molecular profiles from routine histology, potentially enabling scalable and clinically accessible molecular phenotyping.

\begin{figure}[t]
    \centering
    \includegraphics[width=\linewidth]{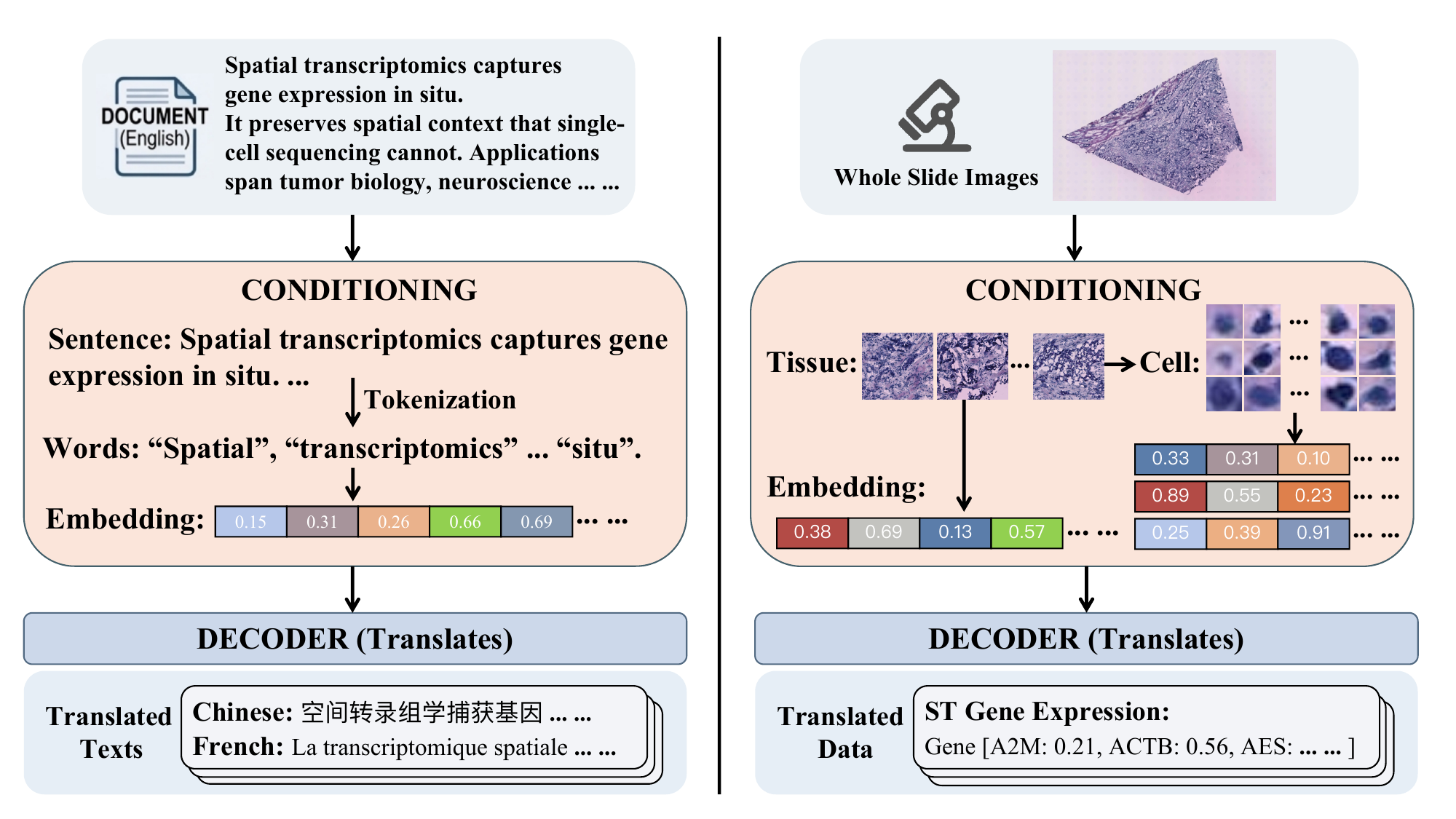}
    \caption{Pathological images exhibit an intrinsic hierarchical structure highly analogous to natural language, framing spatial transcriptomics generation as a translation task facilitates comprehensive modeling of biologically meaningful hierarchical associations, grounding the generation process in authentic biological semantics.}
    \label{fig:figure1}
\end{figure}

Several computational approaches have been developed to infer spatial gene expression from histopathology images~\cite{he2020stnet,Hist2ST,HistoGene,wang2025m2ost,zhu2025stem,ouyang2025genar,chadoutaud2026scellst}. Early methods mainly framed this task as supervised regression, learning direct mappings from image-derived features to spot-level gene expression profiles~\cite{he2020stnet,Hist2ST,HistoGene,monjo2022DeepSpaCE,wang2025m2ost}. More recent studies have explored generative formulations that model gene expression as a conditional distribution given histological context, enabling more flexible characterization of biological heterogeneity and structural organization~\cite{zhu2025stem,ouyang2025genar,huang2025stflow}. Despite these advances, most existing methods still treat the task as a relatively direct image-to-expression prediction problem. This view overlooks two important forms of biological hierarchy. First, spot-level transcriptional states emerge from the joint effects of cellular composition, intrinsic cell states, and the surrounding tissue microenvironment, whereas most current methods primarily rely on generic visual multi-scale representations such as image pyramids, receptive-field expansion, or local-global fusion~\cite{chung2024TRIPLEX,wang2025m2ost,zhang2024iStar}. Second, gene expression is not a flat high-dimensional target, but a structured transcriptional system with coordinated co-expression patterns and cross-gene dependencies~\cite{komili2008coupling,mahat2024single,kunes2024supervised}. Consequently, there remains a need for a biologically grounded framework that explicitly models both the cell-to-tissue semantic hierarchy in histopathology and the hierarchical dependency structure of gene expression.

Gene expression at each spot (a spatial sampling unit) emerges from the interplay between the intrinsic states of its constituent cells and the surrounding spatial microenvironment~\cite{yang2025spotiphy,dong2025simvi}. Histopathological images fundamentally differ from natural images in their intrinsic hierarchical biological organization, a compositional hierarchy akin to that in natural language~\cite{akbar2025CellEcoNet,xiao2025CCI}. As illustrated in Figure~\ref{fig:figure1}, individual cells can be viewed as basic semantic units: their biological meaning is shaped not only by morphology but also by neighboring cells and local tissue context, much like the meaning of a word depends on linguistic context. Multi-cellular patches or spatial spots then function as higher-level semantic constructs analogous to sentences, whose gene expression profiles reflect emergent biological meaning rather than a simple aggregation of cellular features. This analogy suggests that inferring spatial transcriptomic profiles from histopathology is more naturally viewed as a hierarchical cross-modal semantic translation problem than as a direct feature-to-vector regression task.

Inspired by the aforementioned observations, we develop a hierarchical cell–tissue grounded cross-modal semantic translation framework that maps H\&E-stained tissue slides to spatial spot-level gene expression. First, to capture the complex semantics of the source modality and account for the direct regulatory role of cell-type composition on gene expression, we propose a Hierarchical Cell-Tissue Conditioning module. This module fuses cell-type compositional priors and tissue-contextual features via an asymmetric conditioning mechanism that integrates both explicit statistics and implicit cell-type prototypes. An adaptive gating module then dynamically modulates this fusion, ensuring effective semantic alignment between the two feature streams and enabling the model to ground visual representations in biologically meaningful cellular identities. Second, to tackle the intractable high-dimensionality of gene vectors and reflect the modular co-expression structure of transcriptional programs, we employ an autoregressive decoder that decomposes generation into a coarse-to-fine process. This process first models the global expression profile per spot before progressively refining individual gene values. To prevent cross-scale semantic drift during this multi-scale generation, a scale-adaptive mechanism dynamically re-aligns the conditioning signal with the granularity of each generation step. Specifically, we implement this via a scale embedding that biases the conditioning signal to align with the semantic granularity of each generation stage, and a hierarchical FiLM~\cite{perez2018film} that applies progressive modulation from the gene-group level to the cellular microenvironment, preserving biological coherence across scales. Finally, to ensure that the generated profiles possess both high numerical accuracy and biological plausibility, we design a joint supervision objective that enforces three complementary constraints: predictive fidelity, transcriptional distribution statistics, and cell-type semantic structure. This multi-faceted loss guides the model to not only match observed expression values but also respect the underlying statistical and semantic properties of real spatial transcriptomic data.

Our main contributions are summarized as follows:
\begin{enumerate}
\item We propose a novel hierarchical cell–tissue grounded cross-modal translation framework for spatial transcriptomics, wherein cellular semantics and tissue microenvironment are jointly encoded as a unified hierarchical condition.
\item We propose a scale-adaptive autoregressive generation over a predefined hierarchical semantic vocabulary structured by gene co-expression, with dynamic remodulation of semantic conditioning at each granularity level to ensure cross-scale coherence.
\item We propose SpectraLoss, a full-spectrum joint supervision objective that comprehensively constrains gene expression generation across predictive, transcriptional-statistical, and semantic dimensions.
\item Experimental results demonstrate that our framework \\ achieves the state-of-the-art performance on three public benchmarks across diverse species and tissue types.
\end{enumerate}

\begin{figure*}[t]
    \centering
    \includegraphics[width=\linewidth]{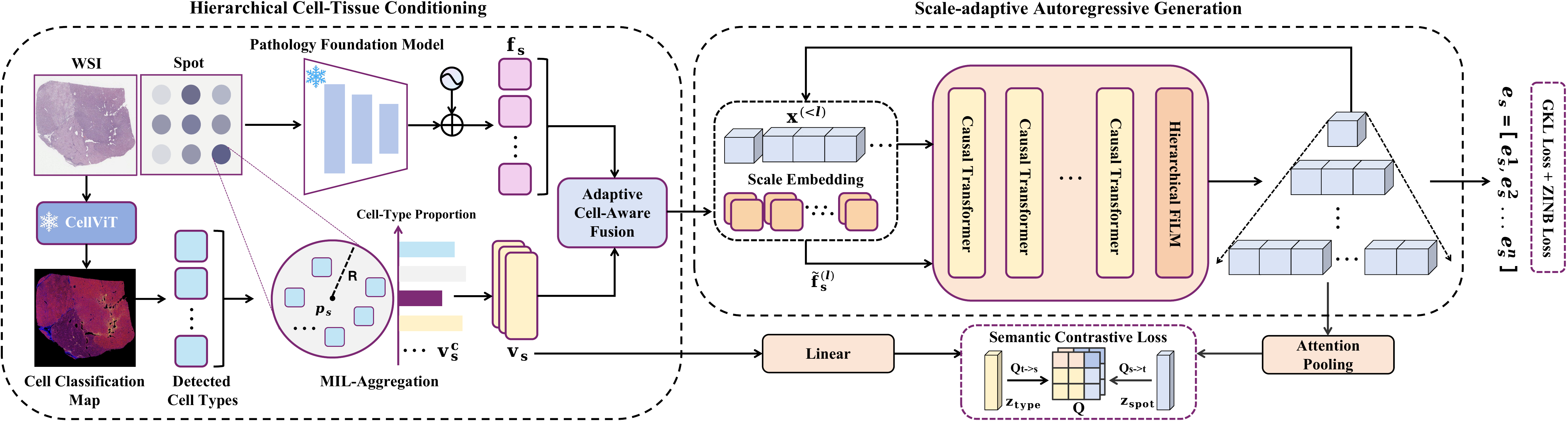}
    \caption{Overview of the Path2ST framework. The proposed framework comprises three key components: Hierarchical Cell-Tissue Conditioning for constructing biologically grounded generation conditions, Scale-Adaptive Autoregressive Generation for coarse-to-fine multi-scale gene expression generation, and SpectraLoss, a composite objective combining GKL, ZINB, and semantic contrastive losses.} 
    \label{fig:figure2}
\end{figure*}

\section{Related Work}

\textbf{Regression-Based Methods.}
Early studies mainly formulated spatial transcriptomic prediction as supervised regression from image features to spot-level gene expression. ST-Net~\cite{he2020stnet} pioneered this direction with DenseNet-based morphological feature extraction and patch-level regression, while DeepSpaCE~\cite{monjo2022DeepSpaCE} incorporated semi-supervised learning to alleviate annotation scarcity. To better capture long-range dependencies, HisToGene~\cite{HistoGene} introduced Vision Transformers, and Hist2ST~\cite{Hist2ST} further combined CNNs, Transformers, and GNNs for multi-level structural modeling. Subsequent methods emphasized higher-order and multi-scale context, including hypergraph-based modeling in HGGEP~\cite{li2024HGGEP} and Ph2st~\cite{niu2025ph2st}, multi-scale tissue encoding in TRIPLEX~\cite{chung2024TRIPLEX} and M2OST~\cite{wang2025m2ost}, and finer-resolution prediction in iStar~\cite{zhang2024iStar} and sCellST~\cite{chadoutaud2026scellst}. However, these methods still largely cast the task as direct image-to-expression regression. In contrast, our method explicitly models the biologically grounded cell--tissue hierarchy and formulates histology-to-transcriptomics inference as structured cross-modal semantic translation.

\textbf{Contrastive Learning-Based Methods.}
Contrastive learning-based approaches align histopathology and transcriptomic signals in a shared latent space instead of directly regressing expression. BLEEP~\cite{xie2023BLEEP} learns bimodal embeddings of H\&E patches and gene expression and predicts through nearest-neighbor retrieval, which also helps mitigate batch effects. NH$^2$2ST~\cite{qu2025NH2ST} further combines dual-scale contrastive learning with hypergraph modeling to enhance cross-modal alignment and spatial interaction modeling. While effective for representation alignment, these methods remain dependent on similarity matching in a shared embedding space and do not explicitly model either cell--tissue semantic hierarchy or structured gene generation. Our framework instead uses unified biological conditioning and hierarchical autoregressive decoding for direct transcriptional generation.

\textbf{Generative Methods.}
Generative models have recently emerged as a promising alternative for spatial transcriptomic prediction. STEM~\cite{zhu2025stem} introduced conditional diffusion modeling to capture expression uncertainty beyond deterministic prediction. STFlow~\cite{huang2025stflow} adopted flow matching to model joint expression distributions across tissue sections. GenAR~\cite{ouyang2025genar} further repurposed next-scale autoregressive generation~\cite{zhou2025next,tian2024visual}, highlighting the value of structured decoding on the transcriptomic side. Beyond these methods, our framework not only performs structured generation of gene expression, but also grounds decoding in an explicitly modeled cell-to-tissue hierarchy on the histopathology side, thereby jointly capturing source-side biological semantics and target-side transcriptional dependencies.

\section{Methodology}
\subsection{Problem Formulation}
Given an H\&E-stained Whole Slide Image (WSI) $\mathcal{I} \in \mathbb{R}^{H \times W \times 3}$, it is typically overlaid with a spatially barcoded array comprising $M$ discrete spots, denoted as $\mathcal{S} = \{1, \ldots, M\}$. Each spot $s$ covers a circular tissue region with a fixed radius, serving as a discrete sampling unit associated with a gene expression profile $\mathbf{e}_s \in \mathbb{N}_0^n$ spanning the gene set $\mathcal{G} = \{1, \ldots, n\}$. The scalar ${e}_s^g$ denotes the mRNA molecule count of gene $g \in \mathcal{G}$ captured at location $s$, representing the expression level of gene $g$ at $s$. The aggregate spatial transcriptomics matrix $\mathbf{X} = [\mathbf{e}_1, \ldots, \mathbf{e}_M]^\top \in \mathbb{N}_0^{M \times n}$ characterizes the global molecular landscape of the tissue. Our goal is to establish a generative mapping $f_\theta: \mathcal{I} \rightarrow \mathbf{X}$ that bridges the modality gap between histological phenotypes and their corresponding spatial molecular profiles.

The hierarchical organization of pathology images exhibits a natural structural analogy to the composition of tokens and sentences in natural language. Consequently, this prediction task can be conceptualized as a cross-modal translation, where the WSI serves as the source language and the spatial transcriptomic profile $\mathbf{X}$ as the target language. Under this formulation, the conditional distribution induced by $f_\theta$ is written as:
\begin{equation}
    p_\theta(\mathbf{X} \mid \mathcal{I}) = \prod_{s=1}^{M} p_\theta(\mathbf{e}_s \mid \mathcal{I}, s).
\end{equation}
This factorization assumes conditional independence across spots, while within each spot the gene expression profile is generated. Inspired by the autoregressive generation paradigm in natural language processing, we explicitly cast the spot-wise cross-modal generative mapping as a conditional autoregressive process.

\subsection{Hierarchical Cell-Tissue Conditioning}
In the conditional autoregressive generation framework, conditioning provides the source information that guides the entire generation process. Unlike existing methods that rely on a single semantic level of histological representation, we propose a conditioning strategy that jointly leverages spot-level histological context and cell-level compositional information. Specifically, explicit conditioning incorporates cell-type proportions, while implicit conditioning adaptively activates learnable cell-type semantics based on local tissue morphology. A dynamic gating mechanism then controls the fusion of cell-tissue information, enabling biologically grounded integration of cellular semantics with histology.

\begin{figure}[t]
    \centering
    \includegraphics[width=\linewidth]{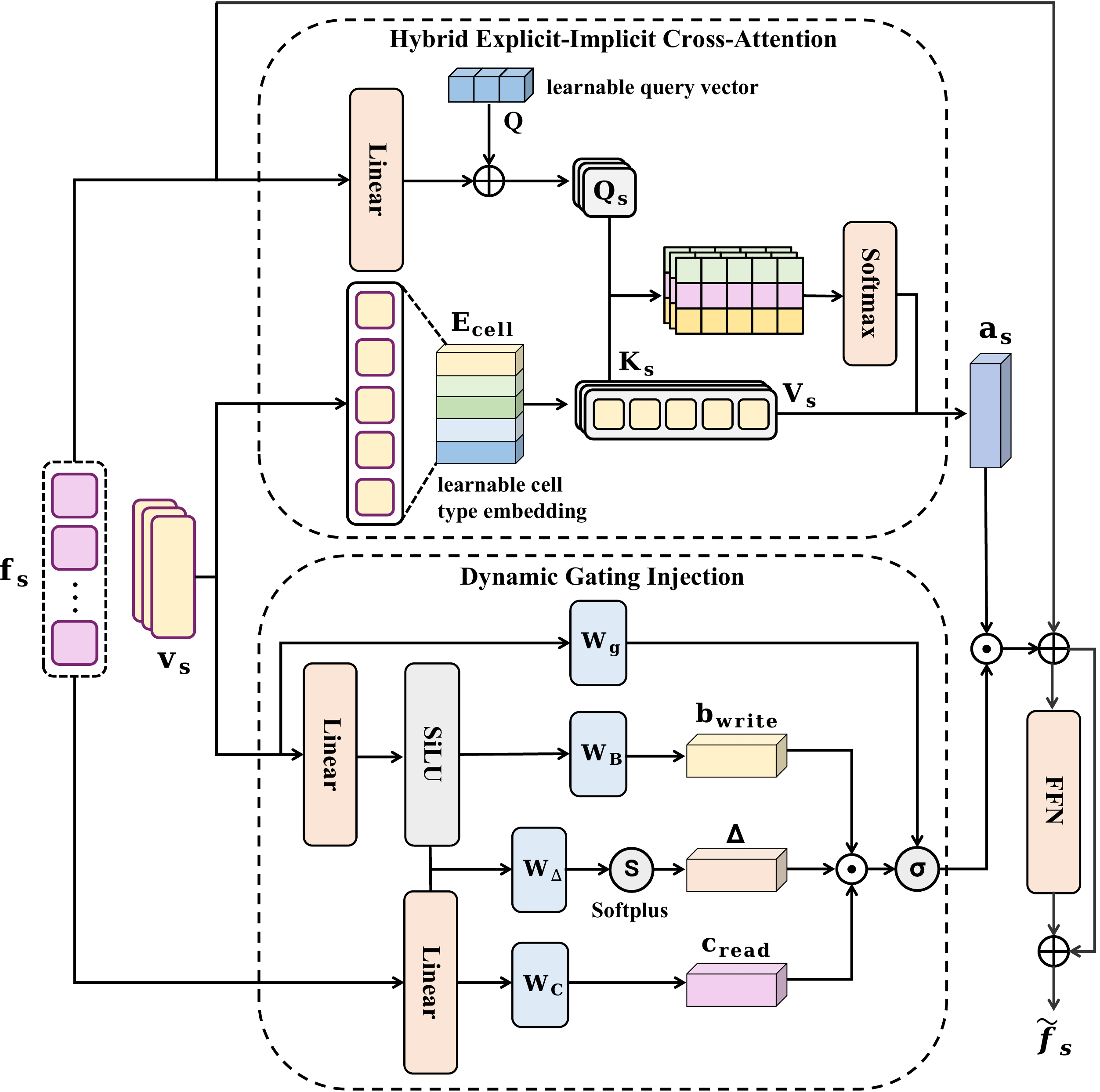}
    \caption{The Hierarchical Cell-Tissue Conditioning module integrates cellular and tissue information via two components: a Hybrid Explicit-Implicit Cross-Attention that captures tissue-context-aware cellular semantic features, and a Dynamic Gating mechanism that adaptively controls the injection strength of fused cellular signals.}
    \label{fig:figure3}
\end{figure}

\textbf{Spatially-aware Histological Embedding.} 
For each spot $s \in \mathcal{S}$, a patch $I_s \in \mathbb{R}^{224 \times 224 \times 3}$ is cropped from $\mathcal{I}$ centred at the coordinate $\mathcal{P}_s \in \mathbb{R}^2$, ensuring complete coverage of the intra-spot cellular and tissue architecture as well as the surrounding tissue microenvironment. The patch $I_s$ is subsequently fed into a pretrained pathology foundation model \cite{chen2024uni} to extract tissue-level histological representations: $\mathbf{t}_s = \phi(I_s) \in \mathbb{R}^{1536}$. The spatial coordinate $\mathcal{P}_s$ is encoded via sinusoidal positional encoding and concatenated with $\mathbf{t}_s$, followed by a linear projection to yield the spatially-aware tissue-level condition embedding $\mathbf{f}_s \in \mathbb{R}^{D}$, which jointly encodes the high-level morphological semantics and spatial context of $s$.

\textbf{MIL-aggregated Cellular Embedding.} To construct cell-level condition embeddings, we first apply a pretrained cell segmentation and classification model~\cite{horst2024cellvit} to the WSI $\mathcal{I}$, yielding the centroid $\mathcal{P}_k \in \mathbb{R}^2$ and cell-type label $y_k \in \{1, \ldots, C\}$ for each detected cell instance $k$, where $C$ is the number of cell types. We then formulate spot-wise multiple instance learning (MIL) by treating each spatial spot $s$ as a bag $\mathcal{B}_s$, comprising all cells whose centroids lie within a radius $R = 112$ pixels of the spot center $\mathcal{P}_s$, matching the cropped patch radius. Formally, $\mathcal{B}_s = \left\{ k \;\middle|\; \|\mathcal{P}_k - \mathcal{P}_s\|_2 \leq R \right\}$.

Cell type reflects fundamental cellular states and intrinsic biological functions, and thus the cell-type composition within bag $\mathcal{B}_s$ strongly shapes the local gene expression profile at spot $s$. To encode this biological prior, we represent each bag by the normalized cell-type proportion vector:
\begin{equation}
    v_s^c = \frac{1}{\max(|\mathcal{B}_s|,1)} \sum_{k \in \mathcal{B}_s} \mathbf{1}[y_k = c], \quad \mathbf{v}_s = [v_s^1, \ldots, v_s^C]^\top \in \mathbb{R}^C
\end{equation}
where $v_s^c$ denotes the fraction of cell type $c$ within bag $\mathcal{B}_s$, and $c \in \{1, \ldots, C\}$. For empty bags ($|\mathcal{B}_s|=0$), we set $\mathbf{v}_s=\mathbf{0}$. This representation preserves biologically relevant semantics while offering a compact and interpretable cell-level condition for downstream generation.

\textbf{Adaptive Explicit--Implicit Cell-Aware Fusion.} 
Given the tissue-level embedding $\mathbf{f}_s$ and the cell-type proportion vector $\mathbf{v}_s$, we design a hybrid explicit--implicit cross-attention module to fuse these signals by selectively attending to cell-type-derived latent signatures most relevant to local gene expression. Specifically, we maintain a shared learnable cell-type embedding matrix $\mathbf{E}_{\mathrm{cell}} \in \mathbb{R}^{C \times D}$, where row $(\mathbf{E}_{\mathrm{cell}})_c$ denotes the semantic prototype of cell type $c$. These prototypes are explicitly weighted by the corresponding proportions in $\mathbf{v}_s$, producing cell-aware key and value tokens:
\begin{equation}
    \mathbf{K}_s = \mathbf{V}_s = \text{diag}(\mathbf{v}_s)\mathbf{E}_{\mathrm{cell}}=\left[ v_s^1 (\mathbf{E}_{\mathrm{cell}})_1,\, \ldots,\, v_s^C (\mathbf{E}_{\mathrm{cell}})_C \right] \in \mathbb{R}^{C \times D}.
\end{equation}
To retrieve expression-relevant cellular semantics, we introduce a set of learnable query vectors $\mathbf{Q} \in \mathbb{R}^{N_q \times D}$, augmented with a spot-specific offset derived from the tissue embedding $\mathbf{f}_s$:
\begin{equation}
    \mathbf{Q}_s = \mathbf{Q} + \delta(\mathbf{f}_s) \in \mathbb{R}^{N_q \times D},
\end{equation}
where $\delta(\cdot)$ is a lightweight projection.

This yields an asymmetric conditional attention architecture: keys and values are explicitly constructed from the cell-type prior $\mathbf{v}_s$, while queries implicitly retrieve the most relevant latent cellular semantics under tissue context. We then apply cross-attention to obtain query-specific cell-aware representations:
\begin{equation}
    \mathbf{A}_s = \text{Softmax}\!\left(\frac{\mathbf{Q}_s\,(\mathbf{K}_s)^\top}{\sqrt{D}}\right)\mathbf{V}_s \in \mathbb{R}^{N_q \times D}
\end{equation}
and average over queries to obtain a single fused cellular representation $\mathbf{a}_s = \frac{1}{N_q}\sum_{j=1}^{N_q}\mathbf{A}_{s,j} \in \mathbb{R}^{D}.$

The resulting $\mathbf{a}_s$ captures cellular semantics conditioned on tissue context. To integrate it adaptively, we introduce a dynamic gating mechanism that modulates the injection strength of $\mathbf{a}_s$ into $\mathbf{f}_s$. Inspired by structured state space models (SSMs)~\cite{gu2024mamba}, the gate is computed from the input rather than fixed priors. Concretely, we project $\mathbf{v}_s$ and $\mathbf{f}_s$ into a low-dimensional space ($r=64$) to obtain write-side and read-side features:
\begin{equation}
    \mathbf{p} = \text{SiLU}(\text{Linear}_{C \to r}(\mathbf{v}_s)) \in \mathbb{R}^{r},
    \quad 
    \mathbf{q} = \text{Linear}_{D \to r}(\mathbf{f}_s) \in \mathbb{R}^{r}, 
\end{equation}
where $\mathbf{p}$ encodes cell-type composition and $\mathbf{q}$ encodes tissue context. Three input-dependent gating components are then computed for dimension-wise modulation:
\begin{equation}
    \Delta = \text{Softplus}(W_{\Delta}[\mathbf{p};\mathbf{q}]), \ 
    \mathbf{b}_{\mathrm{write}} = W_B \mathbf{p}, \ 
    \mathbf{c}_{\mathrm{read}} = W_C \mathbf{q}
\end{equation}
where $\Delta \in \mathbb{R}^{D}$ controls per-dimension injection magnitude, $\mathbf{b}_{\mathrm{write}}\in \mathbb{R}^{D}$ selects dimensions updated by cellular signals, and $\mathbf{c}_{\mathrm{read}}\in \mathbb{R}^{D}$ determines which dimensions are read by tissue context. These terms are combined with a linear offset pathway and passed through a sigmoid to produce the final gate:
\begin{equation}
    \mathbf{g}_s = \sigma\!\left(W_{\text{gate}}\mathbf{v}_s + \Delta \odot \mathbf{b}_{\mathrm{write}} \odot \mathbf{c}_{\mathrm{read}}\right) \in (0,1)^D
\end{equation}
The aggregated vector $\mathbf{a}_s$ is then injected into the tissue embedding via a gated residual connection, followed by FFN and layer normalization:
\begin{equation}
    \tilde{\mathbf{f}}_s = \text{LN}\!\left(\mathbf{f}_s + \mathbf{g}_s \odot \mathbf{a}_s + 
    \text{FFN}(\mathbf{f}_s + \mathbf{g}_s \odot \mathbf{a}_s)\right) \in \mathbb{R}^D
\end{equation}
This design promotes content-aware fusion by enhancing cell--tissue salient dimensions while suppressing incompatible channels. The refined embedding $\tilde{\mathbf{f}}_s$ is used as the conditional signal for the subsequent multi-scale autoregressive generator.

\subsection{Scale-Adaptive Autoregressive Generation}
Motivated by the analogous hierarchical biological organization of pathology images and the co-expression patterns of genes to natural language, we adopt an autoregressive generation paradigm to model inter-gene causal dependencies. However, directly predicting fine-grained gene-specific expression values is prone to error propagation: minor deviations in global semantics are amplified during sequential decoding, degrading fidelity at the gene level. To address this, first we introduce a hierarchical semantic vocabulary over the gene expression space, decomposing generation into a coarse-to-fine conditional scheme. Coarse-grained tokens first establish the global transcriptional state, which then guides progressive refinement toward individual gene expressions. Crucially, each scale receives independent supervision to explicitly constrain cross-scale semantic drift. Second, since autoregressive generation is conditioned on $\tilde{\mathbf{f}}_s$, a scale-invariant conditioning signal cannot adapt to the shifting semantic granularity across scales. We introduce scale-adaptive condition injection in the multi-scale autoregressive generation process, tailoring the conditioning representation to match the resolution of each generation stage.

\textbf{Hierarchical Semantic Vocabulary.} We first construct a multi-scale hierarchical vocabulary over the gene expression space. Given the full gene expression matrix, each gene is represented by its expression vector across all spots. K-means clustering partitions the $n$ genes into $K$ clusters, where genes within a cluster exhibit highly correlated expression profiles, reflecting shared biological functions or co-regulatory mechanisms.

Within each cluster, we build a token hierarchy via recursive mean aggregation. At the coarsest level, all genes in a cluster are pooled into a single token representing the aggregate activity of the functional module. As granularity increases, the cluster is recursively subdivided, and subgroups are mean-pooled until, at the finest scale, each token corresponds to an individual gene’s expression value. During generation, the model produces tokens from coarse to fine: coarse tokens act as global priors that constrain and guide fine-grained prediction, enabling cross-scale consistency while alleviating the burden of high-dimensional direct regression.

\textbf{Multi-scale Autoregressive Generation.} 
For each spot $s$, the model takes the fused representation $\tilde{\mathbf{f}}_s$ as condition input and generates token sequences autoregressively across $L$ scales using an $N$-layer Causal Transformer. Let $\mathbf{x}^{(l)} = [x_1^{(l)}, \ldots, x_{d_l}^{(l)}]$ denote the token sequence at scale $l$, where $d_l$ is the number of tokens and $d_L = n$ at the finest scale. The generation process is factorized as:
\begin{equation}
    p_\theta\!\left(\mathbf{x}^{(1)}, \ldots, \mathbf{x}^{(L)} \mid \tilde{\mathbf{f}}_s\right) = 
    \prod_{l=1}^{L} p_\theta\!\left(\mathbf{x}^{(l)} \mid \mathbf{x}^{(<l)},\ 
    \tilde{\mathbf{f}}_s^{(l)}\right),
\end{equation}
where $\mathbf{x}^{(<l)} = \{\mathbf{x}^{(1)}, \ldots, \mathbf{x}^{(l-1)}\}$ denotes previously generated coarser tokens, which are drawn from ground-truth annotations under teacher forcing during training and from model-generated outputs during inference, and $\tilde{\mathbf{f}}_s^{(l)}$ is the scale-adaptive condition at scale $l$.
The finest-scale output $\mathbf{x}^{(L)} \in V^{d_L}$ consists of discrete token indices from a vocabulary $V$, and the predicted gene expression profile is recovered via a dequantization mapping $\hat{\mathbf{e}}_s = \psi\!\left(\mathbf{x}^{(L)}\right)$.

To match conditioning with semantic granularity, we introduce a learnable scale embedding $\mathbf{b}^{(l)} \in \mathbb{R}^D$ into the gating mechanism:
\begin{equation}
    \mathbf{g}_s^{(l)} = \sigma\!\left(W_\text{gate}\mathbf{v}_s + \Delta \odot \mathbf{b}_{\mathrm{write}} \odot \mathbf{c}_{\mathrm{read}} + 
    \mathbf{b}^{(l)}\right) \in (0,1)^D,
\end{equation}
and compute the scale-adaptive condition as
\begin{equation}
    \tilde{\mathbf{f}}_s^{(l)} = \text{LN}\!\left(\mathbf{f}_s + \mathbf{g}_s^{(l)} \odot 
    \mathbf{a}_s + \text{FFN}\!\left(\mathbf{f}_s + \mathbf{g}_s^{(l)} \odot 
    \mathbf{a}_s\right)\right) \in \mathbb{R}^D,
\end{equation}
which is injected into every Transformer layer via Adaptive Layer Normalization (AdaLN)~\cite{peebles2023adaln} for sample-specific modulation.

After the causal Transformer and output-layer AdaLN, each scale produces hidden states $\mathbf{h}^{(l)} \in \mathbb{R}^{d_l \times D}$, which are further refined by a scale-adaptive hierarchical FiLM before projection to logits.

At coarse scales ($l < L$), each token represents a gene group, but its identity is not explicitly encoded. To provide position-specific semantic attribution, we apply gene-identity FiLM~\cite{perez2018film,ouyang2025genar}. Each position $i$ is assigned a learnable identity embedding $\mathcal{E}_i^{(l)}$, adaptively aligned to resolution $d_l$, from which affine modulation parameters are generated:
\begin{equation}
    \gamma_i^{(l)},\ \beta_i^{(l)} = \mathcal{F}_{\text{id}}\!\left(\mathcal{E}_i^{(l)}\right), \quad 
    \tilde{h}_i^{(l)} = \left(1 + \gamma_i^{(l)}\right) \odot h_i^{(l)} + \beta_i^{(l)}.
\end{equation}
This helps the model maintain awareness of the semantic identity of each gene-group token during cross-scale decoding.

At the finest scale ($l=L$), predicting individual gene expression requires not only gene identity but also spot-specific cellular context. We therefore stack a cell-type FiLM on top of gene-identity modulation, generating spot-level affine parameters from $\mathbf{v}_s$ and broadcasting them across all genes:
\begin{equation}
    \gamma_s^{\text{cell}},\ \beta_s^{\text{cell}} = \mathcal{F}_{\text{cell}}\!\left(\mathbf{v}_s\right), \quad 
    \hat{h}_i^{(L)} = \left(1 + \gamma_s^{\text{cell}}\right) \odot \tilde{h}_i^{(L)} + \beta_s^{\text{cell}}.
\end{equation}
Together, gene-identity and cell-type FiLM form a scale-adaptive hierarchical modulation, progressively refining representations from coarse gene-group semantics to fine-grained, cell-aware expression prediction. Finally, the hidden states are projected to logits and decoded to obtain $\mathbf{x}^{(l)}$, with $\mathbf{x}^{(L)}$ used to derive the final gene expression prediction $\hat{\mathbf{e}}_s$.

\begin{table*}[t]
\centering
\caption{Comparison with state-of-the-art methods across PRAD, HER2ST, and Healthy Mouse Brain datasets. $\uparrow$ indicates higher is better, $\downarrow$ indicates lower is better. Best results are in \textbf{bold}.}
\label{tab:main_results}
\setlength{\tabcolsep}{1.1pt}
\begin{tabular}{l|ccccc|ccccc|ccccc}
\toprule
\multirow{2}{*}{Method} & \multicolumn{5}{c|}{PRAD} & \multicolumn{5}{c|}{HER2ST} & \multicolumn{5}{c}{Healthy Mouse Brain} \\
\cmidrule(lr){2-6} \cmidrule(lr){7-11} \cmidrule(lr){12-16}
 & PCC-10$\uparrow$ & PCC-50$\uparrow$ & PCC-200$\uparrow$ & MSE$\downarrow$ & MAE$\downarrow$
 & PCC-10$\uparrow$ & PCC-50$\uparrow$ & PCC-200$\uparrow$ & MSE$\downarrow$ & MAE$\downarrow$
 & PCC-10$\uparrow$ & PCC-50$\uparrow$ & PCC-200$\uparrow$ & MSE$\downarrow$ & MAE$\downarrow$ \\
\midrule
BLEEP& 0.580 & 0.510 & 0.316 & 2.475 & 1.091 & 0.773 & 0.714 & 0.565 & 1.243 & 0.833 & 0.342 & 0.280 & 0.156 & 1.591 & 0.987 \\
TRIPLEX & 0.620 & 0.544 & 0.423 & 1.319 & 0.836 & 0.783 & 0.714 & 0.586 & 1.212 & 0.857 & 0.501 & 0.445 & 0.312 & 1.157 & 0.822 \\
M2OST  & 0.602 & 0.551 & 0.442 & 1.290 & 0.862 & 0.810 & 0.759 & 0.660 & 1.151 & 0.820 & 0.456 & 0.387 & 0.231 & 1.148 & 0.861 \\
STEM  & 0.636 & 0.555 & 0.403 & 1.457 & 0.857 & 0.831 & 0.770 & 0.625 & 1.199 & 0.787 & 0.526 & 0.452 & 0.331 & 1.235 & 0.864 \\
GenAR & 0.702 & 0.650 & 0.512 & 1.191 & 0.771 & 0.842 & 0.784 & 0.663 & 1.082 & 0.745 & 0.568 & 0.503 & 0.367 & 1.138 & 0.805 \\
\midrule
\textbf{Path2ST} & \textbf{0.767} & \textbf{0.701} & \textbf{0.578} & \textbf{1.005} & \textbf{0.725} &\textbf{0.854} & \textbf{0.801} & \textbf{0.672} & \textbf{0.905} & \textbf{0.717} & \textbf{0.605} & \textbf{0.537} & \textbf{0.414} & \textbf{1.090} & \textbf{0.796} \\
\bottomrule
\end{tabular}
\end{table*}

\subsection{Full-Spectrum Supervision}
Gene expression prediction is a discrete count generation task with sparsity, overdispersion, and long-tailed distributions. Moreover, expression values are tightly coupled with conditional inputs, making it difficult for a single objective to simultaneously ensure predictive accuracy, biological realism, and semantic consistency. To address this, we propose \textit{SpectraLoss}, a multi-level supervision objective with three complementary components: (1) predictive supervision via adaptive Gaussian-target KL divergence, (2) distributional supervision via zero-inflated negative binomial likelihood, and (3) semantic supervision via soft-positive contrastive learning.

\textbf{Adaptive Gaussian-Target KL Divergence Loss.}
We formulate gene expression prediction as ordinal classification over a quantized vocabulary. Instead of one-hot labels, we construct an adaptive Gaussian soft target centered at the ground-truth expression $y = \mathbf{e}_s^g$, with a standard deviation that increases with expression magnitude:
\begin{equation}
    \sigma = \alpha_{\sigma} \cdot y + \beta_{\sigma}, \quad 
    P_\text{gauss}(k|y) = \frac{\exp\!\left(-\dfrac{(k-y)^2}{2\sigma^2}\right)}
    {\sum_{k'}\exp\!\left(-\dfrac{(k'-y)^2}{2\sigma^2}\right)},
\end{equation}
\begin{equation}
    \mathcal{L}_\text{GKL} = \text{KL}\!\left(P_\text{gauss} \,\|\, \text{Softmax}(\text{logits})\right).
\end{equation}
Here, $k \in \{0, 1, \dots, |V|-1\}$ indexes the discrete token vocabulary of size $|V|$, where $V$ covers the full range of observed raw gene expression counts; $k'$ is a dummy variable for normalization, and $\text{logits} \in \mathbb{R}^{|V|}$ denotes the raw scores output by the causal Transformer at the current scale. This loss is applied across all scales to explicitly constrain prediction consistency at different granularities. At the final scale, the Gaussian target is constructed directly from ground-truth hard labels. At intermediate scales, where targets are obtained by mean pooling and thus become continuous, we construct a soft label distribution $P_\text{interp}$ via floor-ceil interpolation and supervise it using the same KL divergence objective.

\textbf{Zero-Inflated Negative Binomial Loss.}
While $\mathcal{L}_\text{GKL}$ supervises predictive deviation, it does not explicitly model the count statistics of gene expression. To capture sparsity, overdispersion, and excess zeros, we introduce the Zero-Inflated Negative Binomial (ZINB) distribution~\cite{eraslan2019zinb} as a distribution-level constraint on the final single-gene predictions. Starting from the continuous output hidden states at the finest scale $\hat{\mathbf{h}}^{(L)} \in \mathbb{R}^{d_L \times D}$, a shared feature extractor predicts the three ZINB parameters for each gene, namely mean $\mu$, dispersion $\theta$, and zero-inflation probability $\pi$, with the negative log-likelihood adopted as the loss:
\begin{equation}
    \mathcal{L}_\text{ZINB} = -\frac{1}{d_L}\sum_g \log p(y_g \mid 
    \mu_g, \theta_g, \pi_g).
\end{equation}
\begin{equation}
    \log p(y \mid \mu, \theta, \pi) = \begin{cases} 
    \log\!\left(\pi + (1-\pi) \cdot \text{NB}(0 \mid \mu, \theta)\right) & y = 0 \\[4pt]
    \log(1-\pi) + \log \text{NB}(y \mid \mu, \theta) & y > 0 
    \end{cases}.
\end{equation}
Here, $\text{NB}(y \mid \mu,\theta)$ denotes the negative binomial distribution parameterized by mean $\mu$ and dispersion $\theta$. The zero-inflation probability $\pi$ models structural zeros, while the negative binomial component captures the mean expression level and dispersion of transcriptional counts. Together, these parameters improve the biological realism and interpretability of generated expression profiles.

\textbf{Soft-Positive Semantic Contrastive Loss.}
The above losses ensure numerical fidelity and count-level plausibility, but they do not explicitly regularize the latent semantic structure. Since spots with similar cell-type compositions should exhibit similar expression patterns, we further introduce a Soft-Positive Semantic Contrastive Loss to align latent representations with cellular composition. Specifically, we use two sources already present in our framework: (1) the final-scale hidden states $\hat{\mathbf{h}}^{(L)} = [\hat{h}_1^{(L)}, \ldots, \hat{h}_{d_L}^{(L)}]^\top \in \mathbb{R}^{d_L \times D}$ produced by the generator, and (2) the input cell-type proportion vectors $\mathbf{v}_s$.

From $\hat{\mathbf{h}}^{(L)}$, we derive a spot-level gene semantic embedding $\mathbf{z}_\text{spot} \in \mathbb{R}^{256}$ via gated attention pooling. Meanwhile, each cell-type proportion vector $\mathbf{v}_s$ is projected into the same space, yielding $\mathbf{z}_\text{type} \in \mathbb{R}^{256}$. Rather than treating all non-matching pairs as hard negatives, we define a soft positive label matrix based on cosine similarities between cell-type compositions: $P_\text{comp}[i,j] = \text{Softmax}_j\!\left(\frac{\cos(\mathbf{v}_i, \mathbf{v}_j)}{\tau_\text{soft}}\right)$.

The predicted similarity distribution is $    Q_{s\to t}[i,j] = \text{Softmax}_j(\mathbf{z}_\text{spot}^i \cdot \mathbf{z}_\text{type}^j \cdot \tau)$,
and the reverse distribution is defined as $Q_{t\to s}[i,j] = \text{Softmax}_j(\mathbf{z}_\text{type}^i \cdot \mathbf{z}_\text{spot}^j \cdot \tau)$.
    
We optimize it using bidirectional KL divergence:
\begin{equation}
    \mathcal{L}_\text{SCL} = \frac{1}{2}\left[\text{KL}(P_\text{comp} \| Q_{s\to t}) + \text{KL}(P_\text{comp}^\top \| Q_{t\to s})\right].
\end{equation}
This objective encourages biologically similar spots to remain close in the latent space and aligns the generative representation with the cell-type condition. The three components jointly form the \textit{SpectraLoss} objective:
\begin{equation}
    \mathcal{L}_\text{Spectra} = \frac{1}{L}\sum_{l=1}^{L}\mathcal{L}^{(l)} + \lambda\,\mathcal{L}_\text{SCL}.
\end{equation}
\begin{equation}
    \mathcal{L}^{(l)} = \begin{cases} 
    \mathcal{L}_\text{GKL}^{(l)} & l < L \\[4pt]
    (1-\alpha_{\text{zinb}})\,\mathcal{L}_\text{GKL}^{(L)} + \alpha_{\text{zinb}}\,\mathcal{L}_\text{ZINB} & l = L.
    \end{cases}
\end{equation}
Together, these three terms provide complementary supervision from predictive, distributional, and semantic perspectives.

\begin{table*}
\centering
\caption{Ablation study on PRAD dataset.}
\label{tab:ablation}
\setlength{\tabcolsep}{6pt}
\begin{tabular}{l|ccccc}
\toprule
Method & PCC-10$\uparrow$ & PCC-50$\uparrow$ & PCC-200$\uparrow$ & MSE$\downarrow$ & MAE$\downarrow$ \\
\midrule
Baseline                        & 0.736 & 0.659 & 0.519 & 1.148 & 0.753 \\
\quad + Hierarchical Conditioning  & 0.748 & 0.673 & 0.549 & 1.080 & 0.738 \\
\quad + Scale-Adaptive Autoregressive Generation            & 0.752 & 0.679 & 0.554 & 1.069 & 0.736 \\
\quad + ZINB Loss                    & 0.756 & 0.686 & 0.558 & 1.058 & 0.733 \\
\quad + Soft-Positive Semantic Contrastive Loss (Ours)          & 0.767 & 0.701 & 0.578 & 1.005 & 0.725 \\
\bottomrule
\end{tabular}
\end{table*}

\section{Experiments}
\subsection{Datasets}
We evaluated our method on three spatial transcriptomics (ST) datasets: \textbf{PRAD}~\cite{erickson2022PRAD}, \textbf{HER2ST}~\cite{andersson2021her2st}, and \textbf{Healthy Mouse Brain}~\cite{vicari2024mouse}. \textbf{PRAD} contains paired ST and histology images from prostatic acinar adenocarcinoma, profiled with the 10x Genomics Visium platform (55\,\textmu m spot diameter, 1,418--4,079 spots per slide), covering benign, transitional, and tumor regions across multiple Gleason grades. \textbf{HER2ST} consists of paired ST and 20$\times$ H\&E-stained histology images from HER2-positive breast cancer, with a 100\,\textmu m spot diameter and 13,594 spots in total, spanning normal, immune-infiltrated, in situ, and invasive carcinoma regions. \textbf{Healthy Mouse Brain} includes paired ST and histology images from healthy mouse brain tissue in the striatum and substantia nigra, profiled with the 10x Genomics Visium platform (55\,\textmu m spot diameter, 2,675--3,617 spots per slide). Together, these datasets cover diverse species, tissues, and disease states, providing a comprehensive benchmark for evaluating model generalization and robustness.

\subsection{Implementation Details}
All experiments were conducted on an NVIDIA A40 GPU. Following prior benchmark protocols~\cite{jaume2024Hest1K,ouyang2025genar}, all patch sizes are set to 224 pixels. The multi-scale autoregressive generation is configured as $(1, 4, 8, 40, 100, 200)$, and we adopt MEND145, SPA148, and NCBI667 as the test sets for the PRAD, HER2ST, and Healthy Mouse Brain datasets, respectively. The model directly predicts raw counts, and $\log_2$ transformation is applied to the predictions at evaluation. Following standard practice~\cite{ouyang2025genar}, the top 200 genes are selected from the intersection of highly expressed and highly variable genes for evaluation. We employ the pre-trained UNI2-h~\cite{chen2024uni} as the pathology foundation model for histopathological feature extraction, and the pre-trained CellViT~\cite{horst2024cellvit} for cell type classification. The classifier categorizes cells into five predefined types: Connective, Neoplastic, Epithelial, Inflammatory, and Necrotic. 

\subsection{Evaluation Metrics}
We evaluate prediction performance using PCC, MSE, and MAE.
\textbf{PCC} measures the correlation between predicted and true expression values of each gene $g$ across all spots: 
\begin{equation}
\text{PCC}_g = \frac{\text{Cov}(\mathbf{E}^g,\, \hat{\mathbf{E}}^g)}{\sqrt{\text{Var}(\mathbf{E}^g)\text{Var}(\hat{\mathbf{E}}^g)}}
\end{equation}
where $\mathbf{E}^g$ and $\hat{\mathbf{E}}^g$ are the true and predicted expression vectors of gene $g$. We report PCC-10, PCC-50, and PCC-200, corresponding to the mean PCC of the top 10, 50, and 200 genes ranked by PCC.

\textbf{MSE} and \textbf{MAE} quantify numerical errors on expression values:
\begin{equation}
    \text{MSE} = \frac{1}{Mn}\sum_{s=1}^{M}\sum_{g=1}^{n}\left(e_s^g - \hat{e}_s^g\right)^2, \quad
    \text{MAE} = \frac{1}{Mn}\sum_{s=1}^{M}\sum_{g=1}^{n}\left|e_s^g - \hat{e}_s^g\right|
\end{equation}
where $e_s^g$ and $\hat{e}_s^g$ denote the true and predicted expression values of gene $g$ at spot $s$.

\begin{table}
\centering
\caption{Hyperparameter analysis on the query number $N_q$ on PRAD dataset.}
\label{tab:query_ablation}
\setlength{\tabcolsep}{6pt}
\begin{tabular}{c|ccccc}
\toprule
Query & PCC-10$\uparrow$ & PCC-50$\uparrow$ & PCC-200$\uparrow$ & MSE$\downarrow$ & MAE$\downarrow$ \\
\midrule
\textbf{1} & \textbf{0.767} & \textbf{0.701} & \textbf{0.578} & \textbf{1.005} & \textbf{0.725} \\
2 & 0.757 & 0.685 & 0.553 & 1.048 & 0.733 \\
3 & 0.751 & 0.671 & 0.535 & 1.069 & 0.765 \\
4 & 0.765 & 0.689 & 0.557 & 1.068 & 0.739 \\
5 & 0.753 & 0.668 & 0.514 & 1.096 & 0.771 \\
\bottomrule
\end{tabular}
\end{table}

\begin{figure*}
    \centering
    \includegraphics[width=\linewidth]{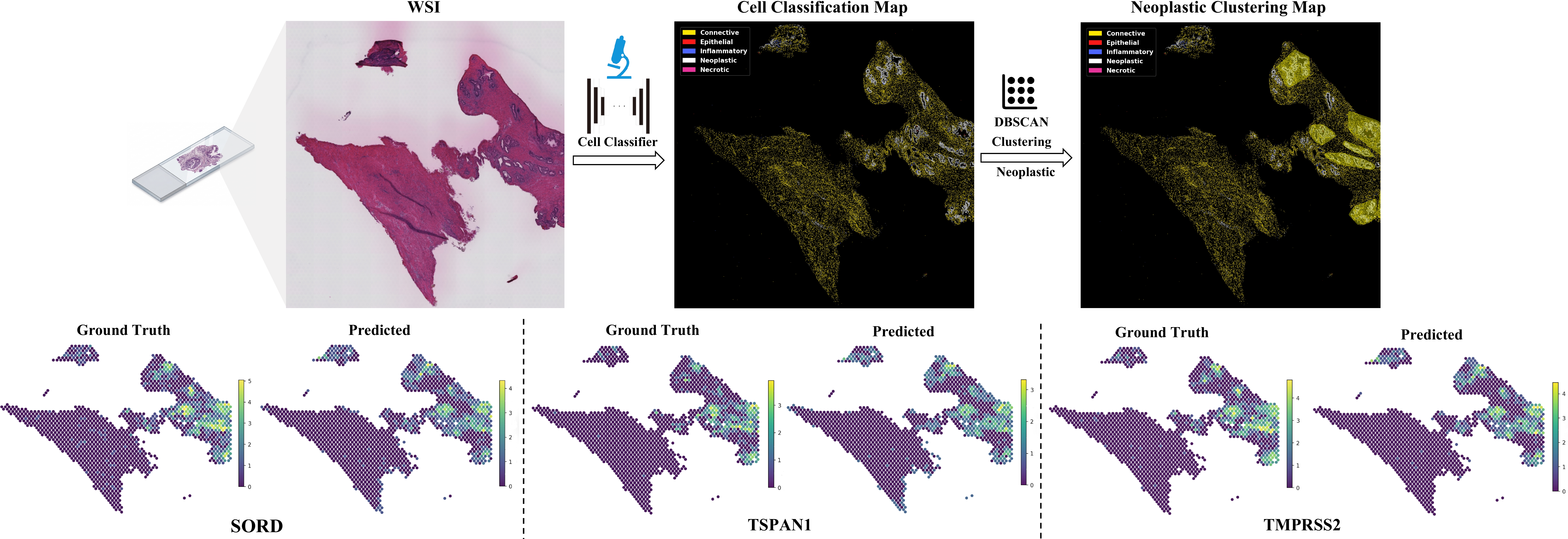}
    \caption{Visualization of cell classification and neoplastic cell clustering on the MEND145 sample from PRAD, alongside spatial expression maps of cancer biomarker genes SORD, TSPAN1, and TMPRSS2, where darker colors indicate lower gene expression at the corresponding spot.}
    \label{fig:vis}
\end{figure*}

\subsection{Comparison with Existing Methods}
We evaluate our framework on PRAD, HER2ST, and Healthy Mouse Brain against existing state-of-the-art methods, including BLEEP (NeurIPS'23)~\cite{xie2023BLEEP}, TRIPLEX (CVPR'24)~\cite{chung2024TRIPLEX}, M2OST (AAAI'25)~\cite{wang2025m2ost}, STEM (ICLR'25)~\cite{zhu2025stem}, and GenAR (MIA'26)~\cite{ouyang2025genar}, covering contrastive, regression, and generative paradigms. As shown in Table~\ref{tab:main_results}, our method achieves the best overall performance across all three benchmarks. On the cancer pathology datasets PRAD and HER2ST, which exhibit high tissue heterogeneity and complex tumour microenvironments, our model consistently outperforms all baselines. The gains are especially notable on the more challenging PRAD dataset: compared with the second-best method GenAR, our model improves PCC-10 and PCC-200 by 6.5\% and 6.6\%, respectively, while reducing MSE and MAE by 0.186 and 0.046. These results indicate superior performance in both highly predictable genes and overall gene expression prediction. On HER2ST, where the performance gap among methods is smaller, our model still achieves consistent improvements over GenAR, with gains of 1.2\% on PCC-10 and 1.7\% on PCC-50.

Subsequently, we evaluate on the healthy brain dataset, which contains a limited number of samples and thus poses additional challenges for model generalisation. Our method still achieves top performance, surpassing the second-best method GenAR by 3.7\% and 4.7\% on PCC-10 and PCC-200, respectively, confirming that our model maintains consistent advantages even under limited data conditions on the dataset of a different species and tissue type.

\subsection{Ablation Study}
We conduct ablation experiments on the PRAD dataset to validate the contribution of each component in our framework, as reported in Table~\ref{tab:ablation}. Starting from the baseline, adding \textit{Hierarchical Cell-Tissue Conditioning} brings a clear performance gain, improving PCC-200 from 0.519 to 0.549 and reducing MSE from 1.148 to 1.080, which confirms the effectiveness of incorporating cellular context into autoregressive conditioning. Further introducing the \textit{Scale-Adaptive Autoregressive Generation} framework yields consistent improvements across all metrics. Adding \textit{ZINB Loss} further enhances count modeling performance, leading to additional gains in both correlation and error metrics. Finally, incorporating the \textit{Soft-Positive Semantic Contrastive Loss} produces the best overall results, increasing PCC-10 to 0.767 and reducing MSE to 1.005. Notably, the largest improvements come from \textit{Hierarchical Cell-Tissue Conditioning} and \textit{Soft-Positive Semantic Contrastive Loss}, highlighting the importance of biologically grounded cellular conditioning and semantic alignment in transcriptomic generation.

\subsection{Hyperparameter Analysis}
In the Hierarchical Cell-Tissue Conditioning module, we adopt an asymmetric design: Keys and Values are explicitly constructed from cell-type proportion vectors to preserve cellular compositional priors, while the Queries are learnable vectors guided by tissue context, adaptively retrieving expression-relevant feature combinations from the cell-type semantic prototype space. To study the effect of query number $N_q$ in this design, we conduct a hyperparameter analysis on the PRAD dataset, as shown in Table~\ref{tab:query_ablation}. The single-query setting achieves the best performance across all metrics. This suggests that, within the explicitly constructed cell-type semantic space, a single query is sufficient to adaptively retrieve the most expression-relevant cellular semantics from a global perspective. Increasing the number of queries leads to consistent performance degradation, likely because multiple queries encourage fragmented associations with local cell-type subspaces and introduce noisy or less relevant signals. These results validate the effectiveness of our asymmetric design and support the use of a single learnable query in the conditioning module.

\subsection{Visualization and Explainability}
Figure~\ref{fig:vis} visualizes the predicted spatial transcriptomics results. The overall spatial expression patterns of the selected genes closely match the ground truth, demonstrating the effectiveness of our method. To further improve biological interpretability, we visualize the cell-type classification results on the WSI, which contains over 60,000 cells. The three selected genes, SORD, TSPAN1, and TMPRSS2, are established prostate cancer biomarker genes associated with disease progression~\cite{lucas2014TMPRSS2,munkley2017TSPAN1,szabo2010SORD}, and all are highly enriched in Neoplastic cells. To verify this spatial association, we apply DBSCAN to the Neoplastic cells identified in the classification map and highlight the resulting high-density regions with yellow masks. In the right tissue region, where Neoplastic cells are densely clustered, all three genes exhibit markedly elevated predicted expression, showing strong spatial concordance with the cell distribution. In particular, the expression peaks align well with the DBSCAN-identified Neoplastic clusters, confirming that higher Neoplastic cell density corresponds to stronger predicted expression of these cancer-related genes. These results support both the biological plausibility of our method and the cell-level interpretability of its predictions.

\subsection{Conclusion}
In this paper, we revisit pathological images from a natural language processing perspective and propose Path2ST, a novel hierarchical cell-tissue grounded cross-modal semantic translation framework that enhances the model's capacity to capture the intrinsic biological hierarchy of pathological images. Through an adaptive explicit-implicit mechanism, Path2ST jointly models the cross-level association between cellular composition and tissue microenvironment, achieving biologically grounded semantic alignment of conditioning signals. The scale-adaptive autoregressive generation further ensures semantic consistency across scales while enabling hierarchical modeling of gene co-expression regulatory relationships. SpectraLoss provides full-spectrum supervision across three complementary dimensions, biological statistical properties, semantic consistency, and numerical fidelity, ensuring that generated profiles are both biologically and statistically meaningful. Extensive experiments demonstrate state-of-the-art performance across multiple benchmarks with strong biological interpretability, highlighting the potential of hierarchical cell-tissue joint modeling for transcriptomic synthesis in digital pathology and offering a cost-effective, accurate, and biologically meaningful solution for routine clinical and research applications.

\begin{acks}
This work was supported by the Zhejiang Provincial Natural Science Foundation of China under Grant No. LQN26F020029.
\end{acks}

\bibliographystyle{ACM-Reference-Format}
\bibliography{reference}

@String{Computing = "Computing" }

@String{Computer = "{IEEE} Computer" }

@article{ouyang2025genar,
  title={GenAR: Next-Scale Autoregressive Generation for Spatial Gene Expression Prediction},
  author={Ouyang, Jiarui and Wang, Yihui and Gao, Yihang and Xu, Yingxue and Yang, Shu and Chen, Hao},
  journal={Medical Image Analysis},
  volume = {114},
  pages = {104232},
  year = {2026}
}

@inproceedings{qu2025NH2ST,
  title={Spatially gene expression prediction using dual-scale contrastive learning},
  author={Qu, Mingcheng and Wu, Yuncong and Di, Donglin and Gao, Yue and Su, Tonghua and Song, Yang and Fan, Lei},
  booktitle={International Conference on Medical Image Computing and Computer-Assisted Intervention},
  pages={574--584},
  year= {2025}
}

@article{he2020stnet,
  title={Integrating spatial gene expression and breast tumour morphology via deep learning},
  author={He, Bryan and Bergenstr{\aa}hle, Ludvig and Stenbeck, Linnea and Abid, Abubakar and Andersson, Alma and Borg, {\AA}ke and Maaskola, Jonas and Lundeberg, Joakim and Zou, James},
  journal={Nature biomedical engineering},
  volume={4},
  number={8},
  pages={827--834},
  year= 2020
}

@inproceedings{HistoGene,
  title={Leveraging information in spatial transcriptomics to predict super-resolution gene expression from histology images in tumors},
  author={Pang, Minxing and Su, Kenong and Li, Mingyao},
  booktitle={BioRxiv},
  doi = {10.1101/2021.11.28.470212},
  year= 2021
}

@article{monjo2022DeepSpaCE,
  title={Efficient prediction of a spatial transcriptomics profile better characterizes breast cancer tissue sections without costly experimentation},
  author={Monjo, Taku and Koido, Masaru and Nagasawa, Satoi and Suzuki, Yutaka and Kamatani, Yoichiro},
  journal={Scientific reports},
  volume={12},
  number={1},
  pages={4133},
  year= 2022
}

@article{Hist2ST,
  title={Spatial transcriptomics prediction from histology jointly through transformer and graph neural networks},
  author={Zeng, Yuansong and Wei, Zhuoyi and Yu, Weijiang and Yin, Rui and Yuan, Yuchen and Li, Bingling and Tang, Zhonghui and Lu, Yutong and Yang, Yuedong},
  journal={Briefings in Bioinformatics},
  volume={23},
  number={5},
  pages={bbac297},
  year= 2022
}

@article{zhang2024iStar,
  title={Inferring super-resolution tissue architecture by integrating spatial transcriptomics with histology},
  author={Zhang, Daiwei and Schroeder, Amelia and Yan, Hanying and Yang, Haochen and Hu, Jian and Lee, Michelle Y.Y. and Cho, Kyung S. and Susztak, Katalin and Xu, George X. and Feldman, Michael D. and others},
  journal={Nature biotechnology},
  volume={42},
  number={9},
  pages={1372--1377},
  year= 2024
}

@article{xie2023BLEEP,
  title={Spatially resolved gene expression prediction from histology images via bi-modal contrastive learning},
  author={Xie, Ronald and Pang, Kuan and Chung, Sai and Perciani, Catia and MacParland, Sonya and Wang, Bo and Bader, Gary},
  journal={Advances in Neural Information Processing Systems},
  volume={36},
  pages={70626--70637},
  year= 2023
}

@inproceedings{chung2024TRIPLEX,
  title={Accurate spatial gene expression prediction by integrating multi-resolution features},
  author={Chung, Youngmin and Ha, Ji Hun and Im, Kyeong Chan and Lee, Joo Sang},
  booktitle={Proceedings of the IEEE/CVF Conference on Computer Vision and Pattern Recognition},
  pages={11591--11600},
  year={2024}
}

@inproceedings{wang2025m2ost,
  title={M2ost: Many-to-one regression for predicting spatial transcriptomics from digital pathology images},
  author={Wang, Hongyi and Du, Xiuju and Liu, Jing and Ouyang, Shuyi and Chen, Yen-Wei and Lin, Lanfen},
  booktitle={Proceedings of the AAAI Conference on Artificial Intelligence},
  volume={39},
  number={7},
  pages={7709--7717},
  year= 2025
}

@article{chadoutaud2026scellst,
  title={sCellST predicts single-cell gene expression from H\& E images},
  author={Chadoutaud, Lo{\"\i}c and Lerousseau, Marvin and Herrero-Saboya, Daniel and Ostermaier, Julian and Fontugne, Jacqueline and Barillot, Emmanuel and Walter, Thomas},
  journal={Nature Communications},
  volume={17},
  number={1},
  pages={1194},
  year= 2026
}

@article{chen2024uni,
  title={Towards a general-purpose foundation model for computational pathology},
  author={Chen, Richard J. and Ding, Tong and Lu, Ming Y. and Williamson, Drew F.K. and Jaume, Guillaume and Song, Andrew H. and Chen, Bowen and Zhang, Andrew and Shao, Daniel and Shaban, Muhammad and others},
  journal={Nature medicine},
  volume={30},
  number={3},
  pages={850--862},
  year={2024}
}

@article{horst2024cellvit,
  title={Cellvit: Vision transformers for precise cell segmentation and classification},
  author={H{\"o}rst, Fabian and Rempe, Moritz and Heine, Lukas and Seibold, Constantin and Keyl, Julius and Baldini, Giulia and Ugurel, Selma and Siveke, Jens and Gr{\"u}nwald, Barbara and Egger, Jan and others},
  journal={Medical image analysis},
  volume={94},
  pages={103143},
  year={2024}
}

@inproceedings{gu2024mamba,
  title={Mamba: Linear-time sequence modeling with selective state spaces},
  author={Gu, Albert and Dao, Tri},
  booktitle={First Conference on Language Modeling},
  year={2024}
}

@inproceedings{perez2018film,
  title={Film: Visual reasoning with a general conditioning layer},
  author={Perez, Ethan and Strub, Florian and De Vries, Harm and Dumoulin, Vincent and Courville, Aaron},
  booktitle={Proceedings of the AAAI conference on artificial intelligence},
  volume={32},
  number={1},
  year={2018}
}

@inproceedings{zhu2025stem,
title={Diffusion Generative Modeling for Spatially Resolved Gene Expression Inference from Histology Images},
author={Sichen Zhu and Yuchen Zhu and Molei Tao and Peng Qiu},
booktitle={The Thirteenth International Conference on Learning Representations},
year={2025},
url={https://openreview.net/forum?id=FtjLUHyZAO}
}

@article{erickson2022PRAD,
  title={Spatially resolved clonal copy number alterations in benign and malignant tissue},
  author={Erickson, Andrew and He, Mengxiao and Berglund, Emelie and Marklund, Maja and Mirzazadeh, Reza and Schultz, Niklas and Kvastad, Linda and Andersson, Alma and Bergenstr{\aa}hle, Ludvig and Bergenstr{\aa}hle, Joseph and others},
  journal={Nature},
  volume={608},
  number={7922},
  pages={360--367},
  year={2022}
}

@inproceedings{jaume2024Hest1K,
  title={Hest-1k: A dataset for spatial transcriptomics and histology image analysis},
  author={Jaume, Guillaume and Doucet, Paul and Song, Andrew H. and Lu, Ming Y. and Almagro-P{\'e}rez, Cristina and Wagner, Sophia J. and Vaidya, Anurag J. and Chen, Richard J. and Williamson, Drew F.K. and Kim, Ahrong and others},
  booktitle={Advances in Neural Information Processing Systems},
  volume={37},
  pages={53798--53833},
  year={2024}
}

@article{andersson2021her2st,
  title={Spatial deconvolution of HER2-positive breast cancer delineates tumor-associated cell type interactions},
  author={Andersson, Alma and Larsson, Ludvig and Stenbeck, Linnea and Salm{\'e}n, Fredrik and Ehinger, Anna and Wu, Sunny Z. and Al-Eryani, Ghamdan and Roden, Daniel and Swarbrick, Alex and Borg, {\AA}ke and others},
  journal={Nature communications},
  volume={12},
  number={1},
  pages={6012},
  year={2021}
}

@article{vicari2024mouse,
  title={Spatial multimodal analysis of transcriptomes and metabolomes in tissues},
  author={Vicari, Marco and Mirzazadeh, Reza and Nilsson, Anna and Shariatgorji, Reza and Bj{\"a}rterot, Patrik and Larsson, Ludvig and Lee, Hower and Nilsson, Mats and Foyer, Julia and Ekvall, Markus and others},
  journal={Nature Biotechnology},
  volume={42},
  number={7},
  pages={1046--1050},
  year={2024}
}

@article{jain2024review,
  title={Spatial transcriptomics in health and disease},
  author={Jain, Sanjay and Eadon, Michael T.},
  journal={Nature reviews nephrology},
  volume={20},
  number={10},
  pages={659--671},
  year={2024}
}

@article{rao2021spatial,
  title={Exploring tissue architecture using spatial transcriptomics},
  author={Rao, Anjali and Barkley, Dalia and Fran{\c{c}}a, Gustavo S. and Yanai, Itai},
  journal={Nature},
  volume={596},
  number={7871},
  pages={211--220},
  year={2021}
}

@article{williams2022introduction,
  title={An introduction to spatial transcriptomics for biomedical research},
  author={Williams, Cameron G. and Lee, Hyun Jae and Asatsuma, Takahiro and Vento-Tormo, Roser and Haque, Ashraful},
  journal={Genome medicine},
  volume={14},
  number={1},
  pages={68},
  year={2022}
}

@article{staahl2016st,
  title={Visualization and analysis of gene expression in tissue sections by spatial transcriptomics},
  author={St{\aa}hl, Patrik L. and Salm{\'e}n, Fredrik and Vickovic, Sanja and Lundmark, Anna and Navarro, Jos{\'e} Fern{\'a}ndez and Magnusson, Jens and Giacomello, Stefania and Asp, Michaela and Westholm, Jakub O. and Huss, Mikael and others},
  journal={Science},
  volume={353},
  number={6294},
  pages={78--82},
  year={2016}
}

@article{schroeder2025scaling,
  title={Scaling up spatial transcriptomics for large-sized tissues: uncovering cellular-level tissue architecture beyond conventional platforms with iSCALE},
  author={Schroeder, Amelia and Loth, Melanie L. and Luo, Chunyu and Yao, Sicong and Yan, Hanying and Zhang, Daiwei and Piya, Sarbottam and Plowey, Edward and Hu, Wenxing and Clemenceau, Jean R. and others},
  journal={Nature methods},
  volume={22},
  number={9},
  pages={1911--1922},
  year={2025}
}

@article{moses2022museum,
  title={Museum of spatial transcriptomics},
  author={Moses, Lambda and Pachter, Lior},
  journal={Nature methods},
  volume={19},
  number={5},
  pages={534--546},
  year={2022}
}

@article{hu2025scalablest,
  title={Scalable spatial transcriptomics through computational array reconstruction},
  author={Hu, Chenlei and Borji, Mehdi and Marrero, Giovanni J. and Kumar, Vipin and Weir, Jackson A. and Kammula, Sachin V. and Macosko, Evan Z. and Chen, Fei},
  journal={Nature biotechnology},
  volume={44},
  number={2},
  pages={215--221},
  year={2026}
}

@article{wang2025benchmarking,
  title={Benchmarking the translational potential of spatial gene expression prediction from histology},
  author={Wang, Chuhan and Chan, Adam S. and Fu, Xiaohang and Ghazanfar, Shila and Kim, Jinman and Patrick, Ellis and Yang, Jean Y.H.},
  journal={Nature Communications},
  volume={16},
  number={1},
  pages={1544},
  year={2025}
}

@article{chelebian2025combining,
  title={Combining spatial transcriptomics with tissue morphology},
  author={Chelebian, Eduard and Avenel, Christophe and W{\"a}hlby, Carolina},
  journal={Nature Communications},
  volume={16},
  number={1},
  pages={4452},
  year={2025}
}

@inproceedings{huang2025stflow,
  title={Scalable Generation of Spatial Transcriptomics from Histology Images via Whole-Slide Flow Matching},
  author={Huang, Tinglin and Liu, Tianyu and Babadi, Mehrtash and Jin, Wengong and Ying, Rex},
  booktitle={Forty-second International Conference on Machine Learning},
  year={2025}
}

@article{yang2025spotiphy,
  title={Spotiphy enables single-cell spatial whole transcriptomics across an entire section},
  author={Yang, Jiyuan and Zheng, Ziqian and Jiao, Yun and Yu, Kaiwen and Bhatara, Sheetal and Yang, Xu and Natarajan, Sivaraman and Zhang, Jiahui and Pan, Qingfei and Easton, John and others},
  journal={Nature Methods},
  volume={22},
  number={4},
  pages={724--736},
  year={2025}
}

@article{dong2025simvi,
  title={SIMVI disentangles intrinsic and spatial-induced cellular states in spatial omics data},
  author={Dong, Mingze and Su, David G. and Kluger, Harriet and Fan, Rong and Kluger, Yuval},
  journal={Nature Communications},
  volume={16},
  number={1},
  pages={2990},
  year={2025}
}

@article{akbar2025CellEcoNet,
  title={Learning the Language of Histopathology Images reveals Prognostic Subgroups in Invasive Lung Adenocarcinoma Patients},
  author={Akbar, Abdul Rehman and Sajjad, Usama and Su, Ziyu and Li, Wencheng and Xing, Fei and Ruiz, Jimmy and Chen, Wei and Niazi, Muhammad Khalid Khan},
  journal={arXiv preprint arXiv:2508.16742},
  year={2025}
}

@article{xiao2025CCI,
  title={Inferring spatial single-cell-level interactions through interpreting cell state and niche correlations learned by self-supervised graph transformer},
  author={Xiao, Xiao and Zhang, Le and Zhao, Hongyu and Wang, Zuoheng},
  journal={Nature Machine Intelligence},
  volume={8},
  number={1},  
  pages={42--58},
  year={2026}
}

@article{li2024HGGEP,
  title={Gene expression prediction from histology images via hypergraph neural networks},
  author={Li, Bo and Zhang, Yong and Wang, Qing and Zhang, Chengyang and Li, Mengran and Wang, Guangyu and Song, Qianqian},
  journal={Briefings in Bioinformatics},
  volume={25},
  number={6},
  pages={bbae500},
  year={2024},
  publisher={Oxford University Press}
}

@article{niu2025ph2st,
  title={Ph2st: St-prompt guided histological hypergraph learning for spatial gene expression prediction},
  author={Niu, Yi and Liu, Jiashuai and Zhan, Yingkang and Shi, Jiangbo and Zhang, Di and Reinius, Marika and Machado, Ines and Crispin-Ortuzar, Mireia and Wu, Jialun and Li, Chen and others},
  journal={arXiv preprint arXiv:2503.16816},
  year={2025}
}

@inproceedings{peebles2023adaln,
  title={Scalable diffusion models with transformers},
  author={Peebles, William and Xie, Saining},
  booktitle={Proceedings of the IEEE/CVF international conference on computer vision},
  pages={4172--4182},
  year={2023}
}

@article{eraslan2019zinb,
  title={Single-cell RNA-seq denoising using a deep count autoencoder},
  author={Eraslan, G{\"o}kcen and Simon, Lukas M. and Mircea, Maria and Mueller, Nikola S. and Theis, Fabian J.},
  journal={Nature communications},
  volume={10},
  number={1},
  pages={390},
  year={2019}
}

@article{mahat2024single,
  title={Single-cell nascent RNA sequencing unveils coordinated global transcription},
  author={Mahat, Dig B. and Tippens, Nathaniel D. and Martin-Rufino, Jorge D. and Waterton, Sean K. and Fu, Jiayu and Blatt, Sarah E. and Sharp, Phillip A.},
  journal={Nature},
  volume={631},
  number={8019},
  pages={216--223},
  year={2024}
}

@article{komili2008coupling,
  title={Coupling and coordination in gene expression processes: a systems biology view},
  author={Komili, Suzanne and Silver, Pamela A.},
  journal={Nature Reviews Genetics},
  volume={9},
  number={1},
  pages={38--48},
  year={2008}
}

@article{kunes2024supervised,
  title={Supervised discovery of interpretable gene programs from single-cell data},
  author={Kunes, Russell Z. and Walle, Thomas and Land, Max and Nawy, Tal and Pe’er, Dana},
  journal={Nature biotechnology},
  volume={42},
  number={7},
  pages={1084--1095},
  year={2024}
}

@article{munkley2017TSPAN1,
  title={The cancer-associated cell migration protein TSPAN1 is under control of androgens and its upregulation increases prostate cancer cell migration},
  author={Munkley, Jennifer and McClurg, Urszula L. and Livermore, Karen E. and Ehrmann, Ingrid and Knight, Bridget and Mccullagh, Paul and Mcgrath, John and Crundwell, Malcolm and Harries, Lorna W. and Leung, Hing Y. and others},
  journal={Scientific reports},
  volume={7},
  number={1},
  pages={5249},
  year={2017}
}

@article{lucas2014TMPRSS2,
  title={The androgen-regulated protease TMPRSS2 activates a proteolytic cascade involving components of the tumor microenvironment and promotes prostate cancer metastasis},
  author={Lucas, Jared M. and Heinlein, Cynthia and Kim, Tom and Hernandez, Susana A. and Malik, Muzdah S. and True, Lawrence D. and Morrissey, Colm and Corey, Eva and Montgomery, Bruce and Mostaghel, Elahe and others},
  journal={Cancer discovery},
  volume={4},
  number={11},
  pages={1310--1325},
  year={2014}
}

@article{szabo2010SORD,
  title={Sorbitol dehydrogenase expression is regulated by androgens in the human prostate},
  author={Szab{\'o}, Zolt{\'a}n and H{\"a}m{\"a}l{\"a}inen, Jenni and Loikkanen, Ildik{\'o} and Moilanen, Anne-Mari and Hirvikoski, Pasi and V{\"a}is{\"a}nen, Timo and Paavonen, Timo K and Vaarala, Markku H.},
  journal={Oncology reports},
  volume={23},
  number={5},
  pages={1233--1239},
  year={2010}
}

@inproceedings{zhou2025next,
  title={Next semantic scale prediction via hierarchical diffusion language models},
  author={Zhou, Cai and Wang, Chenyu and Zhang, Dinghuai and Tong, Shangyuan and Wang, Yifei and Bates, Stephen and Jaakkola, Tommi},
  booktitle={Advances in Neural Information Processing Systems},
  volume = {38},
  pages = {41496--41531},
  year={2026}
}

@article{tian2024visual,
  title={Visual autoregressive modeling: Scalable image generation via next-scale prediction},
  author={Tian, Keyu and Jiang, Yi and Yuan, Zehuan and Peng, Bingyue and Wang, Liwei},
  journal={Advances in neural information processing systems},
  volume={37},
  pages={84839--84865},
  year={2024}
}

\end{document}